\documentclass[runningheads]{llncs}
\usepackage[T1]{fontenc}
\usepackage[pagebackref=true,breaklinks=true,colorlinks,bookmarks=false]{hyperref}
\usepackage{graphicx,verbatim}
\usepackage{subcaption}
\usepackage{comment}
\usepackage{bm}
\usepackage{booktabs}
\usepackage{bold-extra}
\usepackage{xspace}
\usepackage[misc]{ifsym}

\usepackage[table,xcdraw]{xcolor}
\usepackage{dsfont}
\usepackage{colortbl}
\usepackage[frozencache]{minted}
\usepackage{textcomp}
\usepackage{multirow}
\usepackage{algorithm}
\usepackage{algorithmic}
\usepackage{makecell}
\usepackage{amsmath}
\usepackage{amssymb}
\usepackage[figuresright]{rotating}
\usepackage{xspace}

\definecolor{dgreen}{RGB}{0,150,0}

\newif\ifdraft
\drafttrue 

\ifdraft
 \newcommand{\PF}[1]{{\color{red}{\bf PF: #1}}}
 
 \newcommand{\YH}[1]{{\color{dgreen}{\bf YH: #1}}}
 
 \newcommand{\dm}[1]{{\color{violet} #1}}

 \newcommand{\TODO}[1]{\textbf{\color{yellow}[TODO: #1]}}
\else
 \newcommand{\PF}[1]{}
 
 \newcommand{\YH}[1]{}

 \newcommand{\dm}[1]{}
 \newcommand{\TODO}[1]{}
 
\fi

\newcommand{\parag}[1]{\paragraph{#1}}

\newcommand\net{VecHeart}

\begin{document}
\title{\net{}: Holistic Four-Chamber Cardiac Anatomy Modeling via Hybrid VecSets}

\titlerunning{\net{}}

%

\author{Yihong Chen\inst{1} \and Pascal Fua\inst{1}}

\authorrunning{Yihong Chen, Pascal Fua}

\institute{
$^{1}$CVLAB, EPFL \\
\email{\{yihong.chen, pascal.fua\}@epfl.ch}}

\maketitle 


\begin{abstract}

Accurate cardiac anatomy modeling requires the model to be able to handle intricate interrelations among structures. In this paper, we propose VecHeart, a unified framework for holistic reconstruction and generation of four-chamber cardiac structures. To overcome the limitations of current feed-forward implicit methods, specifically their restriction to single-object modeling and their neglect of inter-part correlations, we introduce Hybrid Part Transformer, which leverages part-specific learnable queries and interleaved attention to capture complex inter-chamber dependencies. Furthermore, we propose Anatomical Completion Masking and Modality Alignment strategies, enabling the model to infer complete four-chamber structures from partial, sparse, or noisy observations, even when certain anatomical parts are entirely missing. VecHeart also seamlessly extends to 3D+t dynamic mesh sequence generation, demonstrating exceptional versatility. Experiments show that our method achieves state-of-the-art performance, maintaining high-fidelity reconstruction across diverse challenging scenarios. Code is available at \url{https://github.com/Scalsol/VecHeart}.

 
\end{abstract}


\section{Introduction} 
\label{sec:intro}

Accurate cardiac structure modeling is vital for applications like clinical diagnosis and intervention planning~\cite{sanz2008imaging,li2024solving,qiao2024personalised}. However, faithfully representing multi-part heart geometry from diverse input configurations remains a significant challenge.

Implicit functions offer significant flexibility and have been used across various fields. But conventional implicit-based shape representations~\cite{Park19c,deng2021dif,zheng2021dit,Mescheder19} are typically tailored for single-structure modeling, neglecting the correlations among structures. Recent multi-part approaches~\cite{talabot2026partsdf,yang2024generating,kong2024sdf4chd,liu2024implicit} have begun to address this, but they rely on computationally expensive test-time optimization. Moreover, during test time, they require complete surface information to optimize the latent code. However, the observation is often sparse, and sometimes available only for some structures and not others. For instance, only sparse delineations of the four-chamber structure can be obtained from Long-Axis (LAX) views, while Short-Axis (SAX) MRI stacks cover the left and right ventricles, but not the atrial structures. These constraints severely limit their clinical utility.

To address these limitations, we propose \net{}, a versatile framework for high-quality four-chamber cardiac structure reconstruction and generation given potentially incomplete data. It keeps the fidelity of the individual parts while maintaining overall consistency across the whole heart.

\begin{figure}[t]
    \centering
    \includegraphics[width=\textwidth]{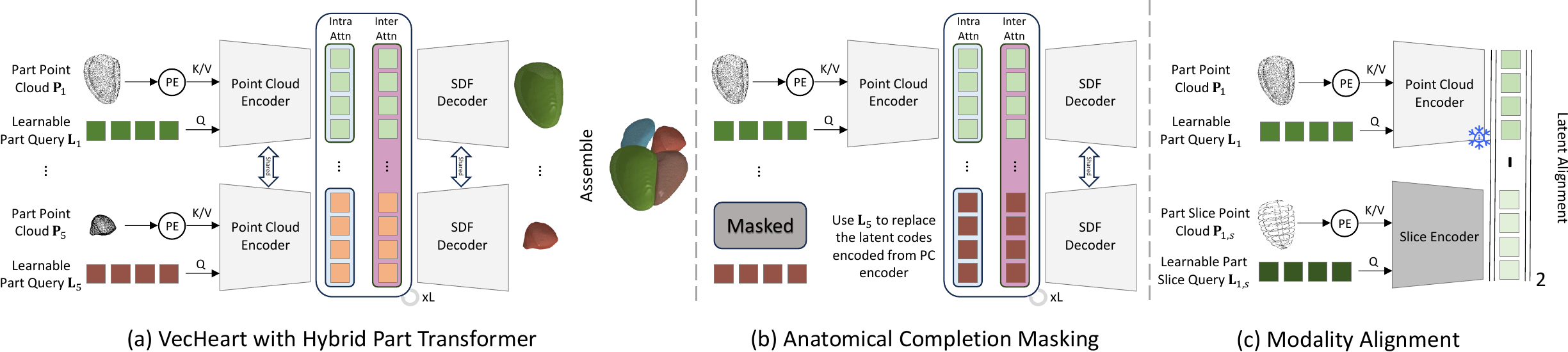}
    \caption{\textit{\net{} pipeline.} (a) A shared encoder creates part latent codes from surface point clouds and learnable part queries that are processed by interleaved intra/inter-part attention layers, which are fed to a shared decoder that outputs SDF values for all parts. (b) During training, we randomly mask $K$ chambers while still making the model reconstruct the whole heart. To this end, we replace the encoded latent vectors by corresponding learnable ones. (c) To handle heterogeneous input, we introduce a dual-encoder and an alignment loss.}    
    \label{fig:pipeline}
\end{figure}

We start from the VecSet~\cite{Zhang23d} latent representation to represent complex surfaces. Given surface points sampled from a 3D shape, VecSet uses a query vector and an encoder to turn them into a set of latent codes that are fed to a set of attention layers and then decoded into a signed distance function. \net{} uses a similar encoder and decoder, along with a part-specific query vector. This enables us to introduce three new techniques designed to handle the heart's complex multi-part anatomy even when the data is missing for some of them: 
\begin{itemize}

 \item \textbf{Hybrid Part Transformer.}  We sequentially and iteratively apply intra-part attention and inter-part attention to capture the correlations between the parts, as shown in Fig.~\ref{fig:pipeline}(a). 
 
 \item \textbf{Anatomical Completion Masking.} During training, we randomly mask some chambers and force \net{} to still output the full anatomy from the incomplete data, as shown by Fig.~\ref{fig:pipeline}(b). This is key to handling missing data.
 
 \item \textbf{Modality Alignment.} To allow reconstruction not only from surface samplings but also from potentially sparse CMR slices as depicted by Fig.~\ref{fig:pipeline}(c), we ensure that codes obtained from such slices are similar to those obtained from denser samplings. 
\end{itemize}
As a result, \net{} provides superior representation ability and a unified solution for robust cardiac modeling across a wide spectrum of input densities, even when some structures are entirely missing. We further show that with minimal effort, we can use flow matching~\cite{Lipman22,esser2024scaling} for 3D+t mesh sequence generation.

\section{Method}


We aim for a consistent representation of the four cardiac chambers, the Left Ventricle (LV), LV myocardium, Right Ventricle (RV), Left Atrium (LA) and Right Atrium (RA). We seek to obtain this multi-component structure from potentially noisy data from different modalities and sometimes missing altogether.

Among the many existing approaches to modeling complex 3D structures, VecSet~\cite{Zhang23d} is considered one of the most powerful. However, it is designed to model individual objects as opposed to multiple anatomically related objects that must remain consistent with each other. Assigning a separate VecSet model to each part would be a simple solution. However, this would not account for their inter-structure relationships. Instead, we introduce \net{}, which retains the encoder and decoder of VecSet but uses different latent vectors for each part along with interleaved attention layers to model inter-part relationships.

We briefly review VecSet~\cite{Zhang23d} in Sec. \ref{sec:preliminary}. In Sec. \ref{sec:vecheart} and \ref{sec:modality}, we introduce \net{}, our proposed framework designed to represent multi-part cardiac structures and handle heterogeneous modalities. Training detail is in \ref{sec:training}. Finally, we describe how \net{} can be extended beyond reconstruction to 3D+t shape generation in Sec. \ref{sec:generation} to showcase the proposed framework's versatility.

\subsection{Preliminary}
\label{sec:preliminary}

VecSet~\cite{Zhang23d} is a transformer-based 3D VAE that encodes surface into compact latent codes. Given a 3D surface $S$, its pipeline consists of three key steps:
\begin{itemize}
    \item \textbf{Surface Sampling}: Sample a point cloud $\mathbf{P}$ of $N$ points from surface $\mathbf{S}$.
    \item \textbf{Feature Encoding}: $\mathbf{P}$ is encoded into compact latent code $\mathbf{C}$ using a mechanism analogous to Perceiver~\cite{jaegle2021perceiver}. This is written as:
    \begin{equation}
        \mathbf{C}=\operatorname{Enc}(\mathbf{L}, \mathbf{P})=\operatorname{Attention}(\mathbf{L}, \operatorname{PosEmb}(\mathbf{P})) \in \mathbb{R}^{D\times M},
    \end{equation}
    where $\operatorname{Attention}(Q, K/V)$ is a standard attention block~\cite{vaswani2017attention}. $\mathbf{L} \in \mathbb{R}^{D\times M}$ is a \textit{learnable query} set with $M$ the set size and $D$ the dimension. 
    \item \textbf{Geometry Decoding}: Given query points $\mathbf{Q}_{space}$ sampled from the space, SDF for $\mathbf{Q}_{space}$ are predicted using a combination of cross/self-attention:
    \begin{equation}
        SDF^{pred}(\mathbf{Q}_{space})=\operatorname{Attention}(\operatorname{PosEmb}(\mathbf{Q}_{space}), \operatorname{Attention}^L(\mathbf{C}, \mathbf{C})),
    \end{equation}
    where the superscript $L$ denotes applying the attention layers $L$ times. 
\end{itemize}
While achieving superior representation capabilities, VecSet is restricted to modeling single object. This renders it unsuitable for directly representing multi-part structures. Assigning a separate VecSet model to each part seems intuitive. However, this will neglect inter-part correlations, which may degrade the performance when dealing with incomplete or sparse input.

\subsection{Structuring Multi-Part Anatomy with \net{}}
\label{sec:vecheart}

To model individual chambers along with myocardium, we introduce the \net{} architecture depicted by Fig. \ref{fig:pipeline}. It expands on VecSet by incorporating a Hybrid Part Transformer (HPT) along with an Anatomical Completion Masking (ACM) strategy. Together, they enable  \net{} to capture inter-part correlations and maintain anatomical integrity, even when the input is incomplete. 

\parag{\bf Hybrid Part Transformer (HPT).}

Given a 4-chamber heart-model $\{\mathbf{S}_p\}_{p=1}^5$ where $\mathbf{S}_p$ denotes the surface of the $p$-th anatomical part, HPT first uses a shared encoder to map each surface into a part-level latent code $\mathbf{C}_p$. To maintain class-wise distinctions while facilitating cross-part information exchange, it relies on \textit{learnable part queries} $\{\mathbf{L}_p\}_{p=1}^5$ to capture part-specific knowledge:
\begin{equation}
    \mathbf{C}_p=\operatorname{Enc}(\mathbf{L}_p, \mathbf{P}_p)=\operatorname{Attention}(\mathbf{L}_p, \operatorname{PosEmb}(\mathbf{P}_p)) \in \mathbb{R}^{D\times M}, p=1,\ldots,5 \;,
\end{equation}
where $\mathbf{P}_p$ is the point cloud sampled from $\mathbf{S}_p$.

Upon obtaining $\{\mathbf{C}_p\}_{p=1}^5$, we facilitate information exchange across components using interleaved intra/inter-part attentions. Specifically, the intra-part attention is applied independently to each $\mathbf{C}_p$ to capture localized features, ensuring that the local anatomical details are preserved and refined. Global inter-part attention is performed over all latent codes $\{\mathbf{C}_p\}_{p=1}^5$ to model structural dependencies and global interactions. Thus we write $\operatorname{IntraAttn}(\mathbf{C}_p)=\operatorname{Attention}(\mathbf{C}_p, \mathbf{C}_p)$ and $ \operatorname{InterAttn}(\mathbf{C}_p)=\operatorname{Attention}(\mathbf{C}_p, \{\mathbf{C}_p\}_{p=1}^5)$.

The SDF for each anatomical part is obtained by querying its part-level latent code through a shared decoder comprising $L$ interleaved attention layers:
\begin{align}
    SDF_p^{pred}(\mathbf{Q}_{space})&=\operatorname{Dec}(\mathbf{Q}_{space}, \mathbf{C}_p), \\
    \operatorname{Dec}(\mathbf{Q}_{space}, \mathbf{C}_p)&=\operatorname{Attention}(\operatorname{PosEmb}(\mathbf{Q}_{space}), \operatorname{XAttn}^L(\mathbf{C}_p)), \nonumber
\end{align}
where $\operatorname{XAttn}=\operatorname{InterAttn}\circ\operatorname{IntraAttn}$, and $\circ$ denotes operation composition.

By using part-specific learnable queries and interleaved attention mechanism, HPT maintains class-wise distinction while facilitating cross-part information exchange. In the following, we detail how HPT can be extended to enable complete four-chamber reconstruction from partial observations when integrated with the Anatomical Completion Masking (ACM) strategy.

\parag{\bf Anatomical Completion Masking (ACM).}

Because data can be missing for some chambers, the ability to perform complete reconstructions from incomplete data is highly desirable. Inspired by Masked Autoencoders (MAE)~\cite{He22a}, we introduce ACM. Specifically, we randomly mask $K$ parts and substitute their extracted latent codes $\mathbf{C}_p$ with the corresponding part-specific learnable queries $\mathbf{L}_p$ as decoder inputs. Through the interleaved attention, the model is tasked with predicting the SDF for \textbf{all} anatomical parts. This encourages the model to infer the missing structures by leveraging both learned anatomical priors and the geometric context provided by other components. We formulate this as:
\begin{align}
    \mathbf{D}_p=(1-\mathbf{M}_p)\mathbf{C}_p+\mathbf{M}_p\mathbf{L}_p \; , \quad
    SDF_p^{pred}(\mathbf{Q}_{space})=\operatorname{Dec}(\mathbf{Q}_{space}, \mathbf{D}_p) \; ,
\end{align}
where $\mathbf{M}_p \in \{0, 1\}$ is a binary mask. $K$ is randomly set to $1$ or $2$ during training. 

\subsection{Modality Alignment (MA)}
\label{sec:modality}

The architecture of Fig.~\ref{fig:pipeline}(b) described above expects point clouds representing the whole chamber as inputs to the encoder. For modalities such as MRI, data is often only available for a few planar slices and reconstruction from such sparse data is a challenge. We address it using latent representations learned from complete surfaces that serve as priors. To this end, we propose our MA strategy.

Let $\{\mathbf{P}_{p, s}\}_{p=1}^5$ be  points from slices corresponding to surfaces $\{\mathbf{S}_p\}_{p=1}^5$. As shown in Fig. \ref{fig:pipeline}(c), we introduce an additional slice encoder $\operatorname{Enc}_{s}$ and learnable slice query sets $\{\mathbf{L}_{p,s}\}_{p=1}^5$ that are used to encode the slice points
\begin{equation}
    \mathbf{C}_{p,s}=\operatorname{Enc}_{s}(\mathbf{L}_{p,s}, \mathbf{P}_{p,s}) \in \mathbb{R}^{D\times M}, p=1,\ldots,5 \; . \label{eq:encoder}
\end{equation}
$\mathbf{C}_{p,s}$ can then be fed to the same frozen decoder as before and the weights of the encoder are learned along the other network weights as described below.

\subsection{Training}
\label{sec:training}

The training of \net{} occurs in two stages. First, we train our model using high-quality complete hearts and minimize the composite loss 
\begin{align}
    \mathcal{L}_{stageI} &=\mathcal{L}_{sdf}+\mathcal{L}_{inter},~ \mbox{with} \;\mathcal{L}_{sdf}=\frac{1}{P}\sum_{p}\mathbb{E}_q\left[||s_p^{pred}(q)-s_p^{gt}(q)||_1\right],\\
    \mbox{and} \;   & \mathcal{L}_{inter} =\sum_{q \in \mathcal{C}}\sum_{p \in \mathcal{N}(q)}\operatorname{tanh}(\max(-\sigma \cdot SDF_{p}^{pred}(q),0)), \nonumber
\end{align}
where $\mathcal{C}$ is the contact regions among parts,  $\mathcal{N}(q)$ are point $q$'s contacting parts, and $\sigma$ a scalar. Minimizing $ \mathcal{L}_{sdf}$ maximizes agreement with the ground-truth while minimizing $\mathcal{L}_{inter}$ discourages intersections. 

Then the model is fine-tuned using SAX and LAX views synthesized as in prior works~\cite{ma2025cardiacflow,xu2023whole,xu2023deep,xu2024improved} by incorporating in-plane motion within SAX stacks and spatial misalignments between SAX and LAX views. These slice displacements are modeled as a Gaussian distribution $\mathcal{N}(0, \lambda^2\mathbf{I})$, where $\lambda$ is sampled from a uniform distribution $\mathcal{U}(0, 6)$ mm. Additionally, we randomly drop LAX slices at a probability $p=0.5$, encouraging the model to predict complete chambers from SAX only slices points. We train the encoder of Eq.~\ref{eq:encoder} by minimizing
\begin{equation}
    \mathcal{L}_{stageII}=\mathcal{L}_{stageI}+\lambda_{la}\mathcal{L}_{la},~~\mathcal{L}_{la}=\frac{1}{P}\sum_{p=1}^P ||\mathbf{C}_{p,s}-\mathbf{C}_{p}||_2,
\end{equation}
where $\mathcal{L}_{la}$ denotes the latent alignment loss. This loss ensures the slice-based encoder inherits rich anatomical priors from the surface-based encoder. Hyperparameters $\lambda_{la}, \sigma$ are set to $0.001, 50$, respectively.

\subsection{Extension to 3D+t Cardiac Shape Generation}
\label{sec:generation}

We demonstrate the versatility of our framework by extending it to generate 3D+t four-chamber mesh sequences. We adopt an architecture similar to Craftsman~\cite{li2024craftsman3d} to jointly generate latent codes $\{C_p\}$ for all anatomical parts. Inspired by CardiacFlow~\cite{ma2025cardiacflow}, we incorporate \textit{Periodic Gaussian Kernel (PGK) encoding}, \textit{Learnable Initial Values}, and \textit{Beta Sampling} to enable one-step generative flow. To handle sequences of different lengths, we apply linear interpolation before using PGK encoding. The model is trained using standard $v$-prediction loss~\cite{esser2024scaling}.

\section{Experiments}

\begin{table}[t]
    \centering
    \caption{\textbf{Heart Shape Reconstruction from Complete/Missing Input.} \textcolor{gray}{Gray} numbers denote the parts are missing from the input. CD in $mm^2$, IoU in \%. P-value: statistical tests on average CD.}
    \label{tab:complete}
    \tabcolsep=2mm
    \renewcommand\arraystretch{0.9}
    \resizebox{1.0\textwidth}{!}{
    \begin{tabular}{lccccccccccc}
        \toprule
        \multirow{2}{*}{Method} & \multicolumn{2}{c}{Myo} & \multicolumn{2}{c}{LV} & \multicolumn{2}{c}{RV} & \multicolumn{2}{c}{LA} & \multicolumn{2}{c}{RA} & \textbf{p-value}\\
        & CD $\downarrow$ & IoU $\uparrow$ & CD $\downarrow$ & IoU $\uparrow$ & CD $\downarrow$ & IoU $\uparrow$ & CD $\downarrow$ & IoU $\uparrow$ & CD $\downarrow$ & IoU $\uparrow$ & (v.s. Ours)\\
        \midrule
        \textit{Complete Input} \\
        ImHeart~\cite{yang2024generating} & 2.085 & 87.19 & 2.085 & 93.86 & 2.809 & 91.61 & 2.308 & 89.53 & 2.363 & 91.07 & <0.01\\
        PartSDF~\cite{talabot2026partsdf} & 0.680 & 92.60 & 0.318 & 96.86 & 0.457 & 95.73 & 0.281 & 95.73 & 0.348 & 95.03 & <0.01\\
        SDF4CHD~\cite{kong2024sdf4chd} & 0.736 & 90.35 & 0.393 & 95.66 & 0.540 & 94.01 & 0.356 & 93.85 & 0.448 & 93.29 & <0.01\\
        CardiacFlow~\cite{ma2025cardiacflow} & 1.609 & 95.97 & 1.115 & 98.09 & 1.591 & 97.46 & 1.396 & 96.92 & 1.451 & 96.84 & <0.01\\
        \midrule
        Ours-Sep & 0.637 & 97.78 & 0.306 & 99.10 & 0.397 & 98.98 & 0.235 & 99.06 & 0.247 & 99.02 & 0.0392\\
        Ours & \textbf{0.617} & \textbf{97.95} & \textbf{0.283} & \textbf{99.16} & \textbf{0.378} & \textbf{99.00} & \textbf{0.215} & \textbf{99.11} & \textbf{0.236} & \textbf{99.05} & -\\
        \midrule
        \textit{Missing Part} \\
        \multirow{2}{*}{ImHeart~\cite{yang2024generating}} & 2.113 & 86.79 & 2.164 & 92.72 & 2.894 & 91.26 & \color{gray}{9.584} & \color{gray}{81.47} & 2.432 & 90.13 & <0.01\\
        & 2.115 & 86.32 & 2.307 & 92.47 & 2.853 & 90.87 & 2.337 & 88.99 & \color{gray}{9.766} & \color{gray}{85.65} & <0.01\\
        \multirow{2}{*}{PartSDF~\cite{talabot2026partsdf}} & 0.688 & 92.46 & 0.327 & 96.60 & 0.455 & 95.69 & \color{gray}{13.180} & \color{gray}{79.38} & 0.355 & 94.89 & <0.01\\
        & 0.686 & 92.53 & 0.326 & 96.68 & 0.455 & 95.65 & 0.287 & 95.60 & \color{gray}{11.536} & \color{gray}{83.83} & <0.01\\
        \multirow{2}{*}{SDF4CHD~\cite{kong2024sdf4chd}} & 0.737 & 90.13 & 0.408 & 95.24 & 0.532 & 93.54 & \color{gray}{4.268} & \color{gray}{88.80} & 0.457 & 92.86 & <0.01\\
        & 0.739 & 90.16 & 0.407 & 95.23 & 0.531 & 93.55 & 0.364 & 93.16 & \color{gray}{5.061} & \color{gray}{88.88} & <0.01 \\
        \multirow{2}{*}{CardiacFlow~\cite{ma2025cardiacflow}} & 1.638 & 94.42 & 1.173 & 97.56 & 1.723 & 96.72 & \color{gray}{5.679} & \color{gray}{88.70} & 1.513 & 95.57 & <0.01 \\
        & 1.665 & 94.30 & 1.175 & 97.50 & 1.696 & 96.75 & 1.447 & 95.90 & \color{gray}{5.956} & \color{gray}{88.22} & <0.01 \\
        \midrule
        \multirow{3}{*}{Ours} & \textbf{0.626} & \textbf{97.65} & \textbf{0.290} & \textbf{98.99} & \textbf{0.386} & \textbf{98.82} & \textbf{\color{gray}{1.938}} & \textbf{\color{gray}{92.89}} & \textbf{0.246} & \textbf{98.82} & - \\
        & \textbf{0.626} & \textbf{97.65} & \textbf{0.289} & \textbf{99.02} & \textbf{0.387} & \textbf{98.79} & \textbf{0.212} & \textbf{98.94} & \textbf{\color{gray}{1.229}} & \textbf{\color{gray}{94.18}} &-  \\
        & 0.640 & 97.20 & 0.301 & 98.81 & 0.403 & 98.52 & \color{gray}{5.979} & \color{gray}{87.03} & \color{gray}{4.336} & \color{gray}{89.48} & - \\
        \bottomrule
    \end{tabular}
    }
\end{table}

To train and test \net{}, we curated a high-quality cardiac mesh dataset comprising 1,060 samples, generated from expert annotations in publicly available datasets, including LAA~\cite{zeng2023imagecas,hansen2025laa}, WHS++~\cite{zhuang2019mmwhs}, and other benchmarks~\cite{metz2009coronary,schaap2009standardized,kiricsli2013standardized,tobon2015benchmark}. Each four-chamber mesh is rigidly registered to a WHS atlas~\cite{zhuang2016multi}, with a 6:2:2 train/val/test split. For reconstruction from real slices and 3D+t mesh sequence generation, we use a dataset comprises 835 CMR sequences constructed from three publicly available MRI datasets: {\it ACDC}, {\it M\&Ms}, and {\it M\&Ms-2}~\cite{bernard2018acdc,campello2021mm1,martin2023mm2}. As in earlier studies~\cite{Meng23b,yuan2023myo4d,xiao2024slice2mesh,qiao2024personalised}, we register and fit a Statistic Shape Model~\cite{duan2019automatic,bai2015bi} to segmentation masks predicted by a publicly available method~\cite{bai2018automated} to generate ground-truth cardiac meshes. We only evaluate LV myocardium, LV and RV as this dataset mainly contain SAX stacks. We use standard L2 Chamfer Distance (CD) and IoU score at resolution $128$ to measure 3D reconstruction quality.

In the \net{} model of Fig.~\ref{fig:pipeline}, the decoder comprises $L=3$ interleaved attention layers. For the part-specific learnable queries $\mathbf{L}_p \in \mathbb{R}^{D \times M}$, the set size $M$ is 256 and dimension $D$ is 32. We train the model for $1000/500$ epochs for stage I/II using Adam~\cite{Kingma14}. All experiments are conducted using one V100 GPU.

\subsection{Heart Shape Modeling for Complete/Incomplete Data}

In this section, we evaluate the shape modeling ability for complete/incomplete data. We use the same setup as conventional approaches~\cite{yang2024generating,kong2024sdf4chd} to reconstruct the test (unknown) samples. We compare against three multi-part auto-decoding implicit methods, PartSDF~\cite{talabot2026partsdf}, ImHeart~\cite{yang2024generating}, SDF4CHD~\cite{kong2024sdf4chd}, as well as with CardiacFlow~\cite{ma2025cardiacflow}, an auto-encoding voxel-based method. For fairness, we train CardiacFlow using the same ACM training strategy as us. For completeness, we also evaluate \textit{Ours-Sep}, that is, fitting a separate VecSet model to each part. 


\begin{figure}[t]
    \centering
    \includegraphics[width=0.9\linewidth]{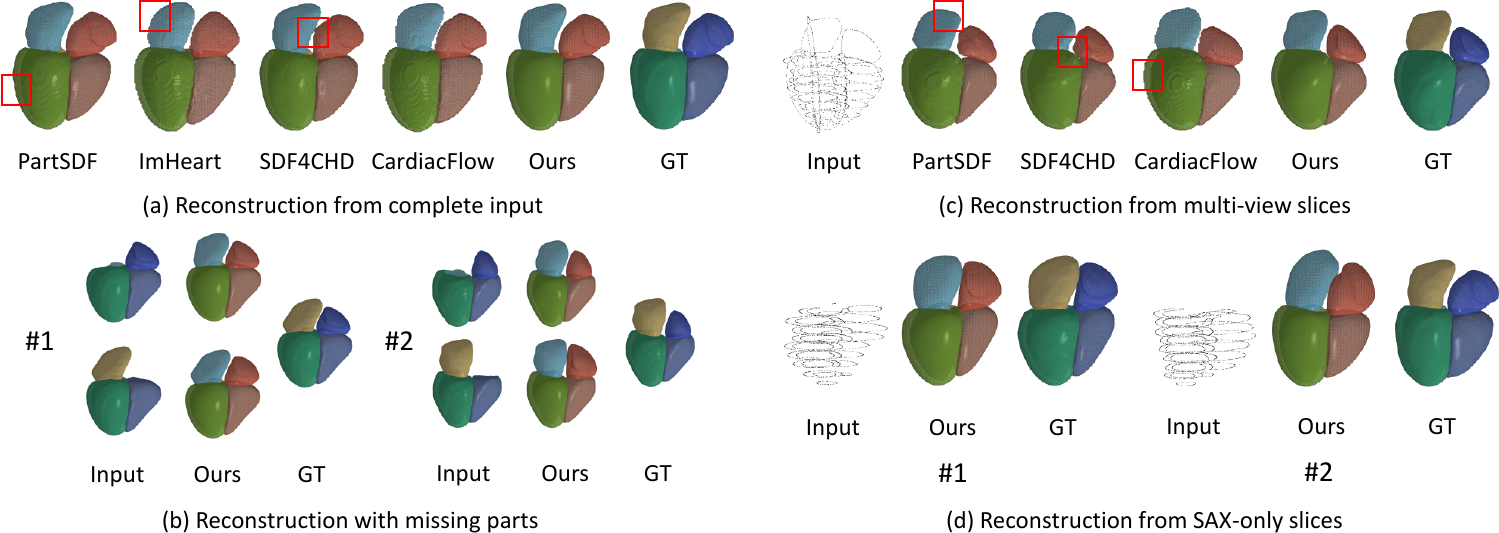}
    \caption{Reconstruction results from diverse input configurations.}
    \label{fig:recon}
\end{figure}

\begin{figure}[t]
     \centering
     \begin{minipage}{0.43\textwidth}
         \centering
         \includegraphics[width=\textwidth]{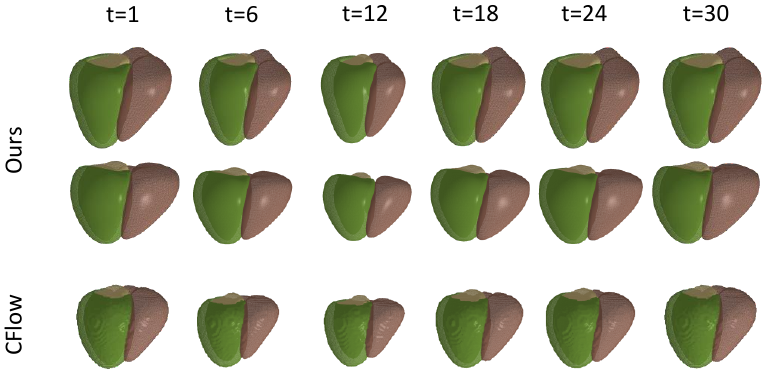}
         \caption{3D+t cardiac shape generation results comparison.}
         \label{fig:seq_vis}
      \end{minipage}
      \hfill
      \begin{minipage}{0.52\textwidth}
          \centering
          \includegraphics[width=\textwidth]{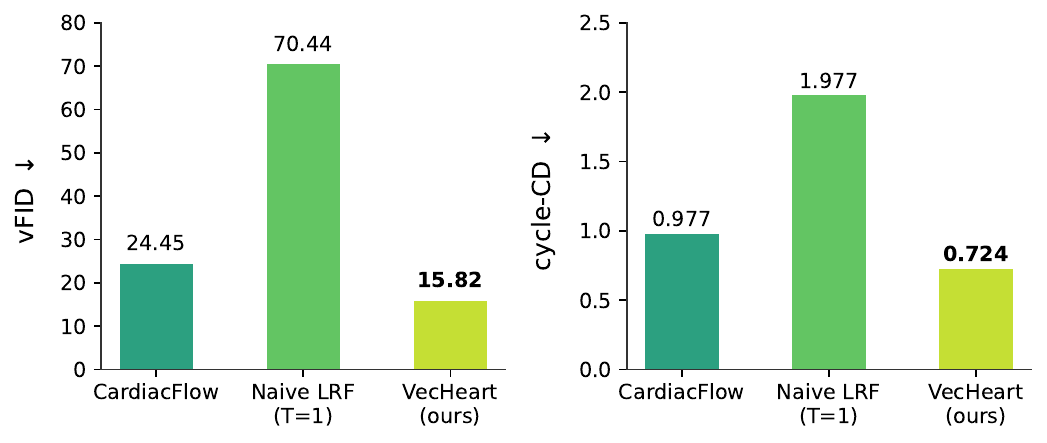}
          \caption{3D+t cardiac shape generation evaluated by vFID and cycle-CD.}
          \label{fig:seq_histo}
      \end{minipage}
\end{figure}

We present qualitative results in Fig. \ref{fig:recon}(a)(b) and report quantitative ones in Tab.~\ref{tab:complete}. The gray numbers are CD and IOU for parts which no data was provided and whose shape was therefore inferred from other parts. To obtain these numbers for auto-decoding methods, we compute the loss on available components and minimize it with respect to the latent vectors.

VecHeart outperforms the baselines in both complete and missing data scenarios. ImHeart's lower accuracy likely stems from its occupancy-based representation. While implicit methods like PartSDF deliver respectable results, they are computationally expensive (e.g., 80s vs. our 0.7s feed-forward inference). While CardiacFlow yields the second-best performance on IoU, it is subject to resolution and discretization constraints. Compared to Ours-Sep, which fail to model inter-part relationships and cannot extend to missing-input scenarios, our full model shows the necessity of modeling inter-part correlation and structural priors. If only HPT is used without the ACM strategy, \net{} cannot produce any meaningful results. These results validate our HPT and ACM designs.

\subsection{Shape Reconstruction from Sparse Slices}

We now turn to sparse and noisy CMR slices. CardiacFlow adopts Label Completion U-Net~\cite{Ronneberger15,xu2023whole} as in the original work to recover full volumes from sparse slices. We experimented with both synthetic and real data. In the synthetic case, following CardiacFlow, we synthesized 10 different multi-view 2D slices at random motion level $\lambda$ using the displacement strategy of Sec. \ref{sec:training} for each test sample. For real experiments, we use the SAX contours of the CMR dataset.

As shown in Fig. \ref{fig:recon}(c)(d), VecHeart produces plausible four-chamber reconstructions from sparse slices, underscoring the practical utility and robustness of our approach. We report quantitative results for synthetic data in the top rows of Tab.~\ref{tab:slice} and for real data in the bottom. As before, the gray numbers denote results for structures which is absent from the input. Our model consistently outperforms the baselines in all configurations. We also compare against {\it Ours-SliceOnly}, which uses the same slice encoder and decoder as the full model but without Stage I training. It works less well, thus confirming the importance of the proposed Modality Alignment strategy. 

Note that the results on real-world CMR data were obtained by training on synthetic data and without re-training. They can be further improved by fine-tuning on real data, as shown by the numbers corresponding to \textit{Ours-Finetune}. 

\begin{table}[t]
    \centering
    \caption{\textbf{Shape Reconstruction from Sparse Slices.} \textcolor{gray}{Gray} numbers denote results for parts for which no input data is supplied. CD in $mm^2$, IoU in \%.}
    \label{tab:slice}
    \tabcolsep=2.0mm
    \renewcommand\arraystretch{0.9}
    \resizebox{1.0\textwidth}{!}{
    \begin{tabular}{lccccccccccc}
        \toprule
        \multirow{2}{*}{Method} & \multicolumn{2}{c}{Myo} & \multicolumn{2}{c}{LV} & \multicolumn{2}{c}{RV} & \multicolumn{2}{c}{LA} & \multicolumn{2}{c}{RA} & \textbf{p-value}\\
        & CD $\downarrow$ & IoU $\uparrow$ & CD $\downarrow$ & IoU $\uparrow$ & CD $\downarrow$ & IoU $\uparrow$ & CD $\downarrow$ & IoU $\uparrow$ & CD $\downarrow$ & IoU $\uparrow$ & (v.s. Ours)\\
        
        \midrule
        \textit{Synthetic SAX, 2CH, 4CH} \\
        PartSDF~\cite{talabot2026partsdf} & 2.434 & 80.74 & 2.663 & 86.35 & 2.927 & 86.28 & 3.034 & 85.50 & 8.565 & 79.97 & <0.01\\
        SDF4CHD~\cite{kong2024sdf4chd} & 1.810 & 82.39 & 1.369 & 91.80 & 2.058 & 89.68 & 2.881 & 88.85 & 6.686 & 82.82 & <0.01 \\
        CardiacFlow~\cite{ma2025cardiacflow} & 2.411 & 86.42 & 1.728 & 91.12 & 2.458 & 88.43 & 2.734 & 87.03 & 4.997 & 83.01 & <0.01\\
        \midrule
        Ours-SliceOnly & 1.430 & 85.02 & 1.091 & 91.10 & 1.456 & 89.86 & 1.481 & 89.26 & 2.977 & 86.32 & <0.01\\
        Ours & \textbf{1.130} & \textbf{88.39} & \textbf{0.732} & \textbf{91.92} & \textbf{1.114} & \textbf{90.25} & \textbf{0.854} & \textbf{90.80} & \textbf{1.571} & \textbf{89.66} & -\\
        
        \midrule
        \textit{Synthetic SAX} \\
        PartSDF~\cite{talabot2026partsdf} & 4.066 & 73.37 & 4.452 & 83.07 & 3.956 & 84.46 & \color{gray}{37.970} & \color{gray}{65.36} & \color{gray}{47.784} & \color{gray}{61.34} & <0.01\\
        SDF4CHD~\cite{kong2024sdf4chd} & 2.676 & 79.05 & 2.445 & 89.83 & 2.624 & 88.62 & \color{gray}{19.822} & \color{gray}{69.56} & \color{gray}{18.568} & \color{gray}{68.64} & <0.01 \\
        CardiacFlow~\cite{ma2025cardiacflow} & 2.591 & 83.58 & 2.606 & 88.35 & 2.709 & 86.13 & \color{gray}{14.207} & \color{gray}{71.33} & \color{gray}{11.898} & \color{gray}{72.95} & <0.01\\
        \midrule
        Ours-SliceOnly & 2.172 & 82.02 & 1.834 & 88.02 & 2.427 & 84.12 & \color{gray}{14.625} & \color{gray}{73.73} & \color{gray}{11.310} & \color{gray}{74.99} & 0.0141\\
        Ours & \textbf{1.816} & \textbf{84.31} & \textbf{1.476} & \textbf{89.26} & \textbf{2.013} & \textbf{87.34} & \textbf{\color{gray}{9.224}} & \textbf{\color{gray}{76.40}} & \textbf{\color{gray}{7.909}} & \textbf{\color{gray}{78.72}} & -\\
        
        \midrule
        \midrule
        \textit{Real SAX (Zero-shot)} \\
        PartSDF~\cite{talabot2026partsdf} & 4.721 & 74.24 & 4.212 & 77.50 & 6.482 & 70.42 & - & - & - & - & <0.01\\
        SDF4CHD~\cite{kong2024sdf4chd} & 3.653 & 77.25 & 3.281 & 80.64 & 5.233 & 73.94 & - & - & - & - & <0.01 \\
        CardiacFlow~\cite{ma2025cardiacflow} & 3.168 & 79.17 & 2.714 & 81.05 & 4.137 & 76.34 & - & - & - & - & <0.01\\
        Ours & \textbf{2.471} & \textbf{79.87} & \textbf{2.296} & \textbf{81.44} & \textbf{3.621} & \textbf{78.10} & - & - & - & - & -\\
        \midrule
        Ours-Finetune & 2.108 & 83.59 & 1.757 & 84.57 & 3.152 & 82.19 & - & - & - & - & -\\  
        \bottomrule
    \end{tabular}
    }
\end{table}

\subsection{3D+t Cardiac Shape Generation}
Now we evaluate VecHeart's capability to generate 3D+t cardiac sequences. As in CardiacFlow, we employ the Volume Fréchet Inception Distance (vFID) and Cycle-CD metrics to assess the diversity and periodic consistency of the generated sequences \cite{heusel2017gans}. Fig. \ref{fig:seq_vis}, \ref{fig:seq_histo} shows we achieve superior qualitative and quantitative performance, producing diverse and high-fidelity sequences. As a contrast, CardiacFlow suffers from discretization artifacts. The results imply \net{}'s potential to be integrated with clinical metadata for more controllable and interpretable generation and become a general purpose data augmenter.


\section{Conclusion}

We present \net{}, a unified framework for holistic reconstruction and generation of four-chamber cardiac anatomy. It delivers high-fidelity reconstruction across a wide spectrum of input densities ranging from dense point clouds to sparse and noisy slices. Furthermore, integrating a flow-matching scheme enables the \net{} to generate realistic 3D+t mesh sequences.

\net{}'s flexibility makes it possible to use partially labeled data, giving it the potential for cross-modality data fusion and robust clinical diagnostics in heterogeneous real-world scenarios. Moving forward, we aim to use \net{} to reconstruct cardiac anatomy from more modalities, including raw MRI and US images acquired over time, within a unified framework.

\begin{credits}
\subsubsection{\ackname} This work was supported by a SNSF grant.
\subsubsection{\discintname} All authors have no conflict of interest.
\end{credits}

\bibliographystyle{splncs04}
\bibliography{bib/short,bib/vision,bib/graphics,bib/biomed,bib/learning,main}

\end{document}